%% file: main.tex
\documentclass[letterpaper, 10 pt, conference]{ieeeconf}  %

\IEEEoverridecommandlockouts                              %

\usepackage{amsmath,amssymb,amsfonts}
\usepackage{graphicx}
\usepackage{hyperref}
\usepackage[capitalize]{cleveref}
\usepackage{listings}
\usepackage{makecell}
\usepackage{optidef}
\usepackage{tikz}
\let\labelindent\relax
\usepackage{enumitem}
\usepackage{subcaption}
\usepackage{xspace}
\usepackage{paralist}
\usepackage{makecell}
\usepackage{subcaption}
\usepackage{tabularx}

\usetikzlibrary{arrows.meta, automata, shapes, positioning, fit, calc, patterns.meta, tikzmark}

\input{macros}

\title{\LARGE \bf
Context-Triggered Contingency Games\\ for Strategic Multi-Agent Interaction
}

\author{Kilian Schweppe$^{1}$ and Anne-Kathrin Schmuck$^{1}$%
\thanks{$^{1}$Max Planck Institute for Software Systems, Kaiserslautern, Germany
        {\tt\small \{kschweppe,akschmuck\}@mpi-sws.org}}%
}

\newtheorem{example}{Example}

\begin{document}

\maketitle
\thispagestyle{empty}
\pagestyle{empty}

\begin{abstract}
We address the challenge of reliable and efficient interaction in autonomous multi-agent systems, where agents must balance long-term strategic objectives with short-term dynamic adaptation. We propose context-triggered contingency games, a novel integration of strategic games derived from temporal logic specifications with dynamic contingency games solved in real-time. Our two-layered architecture leverages strategy templates to guarantee satisfaction of high-level objectives, while a new factor-graph–based solver enables scalable, real-time model predictive control of dynamic interactions. The resulting framework ensures both safety and progress in uncertain, interactive environments. We validate our approach through simulations and hardware experiments in autonomous driving and robotic navigation, demonstrating efficient, reliable, and adaptive multi-agent interaction.
\end{abstract}

\section{Introduction}\label{sec:Intro}
\input{sections/intro}

\section{Problem Setup}\label{sec:Setup}
\input{sections/problemSetup}

\section{Combining Strategic \& Dynamic Games}\label{sec:Games}
\input{sections/combining}

\section{Solving Contingency Games Efficiently}\label{sec:Solver}
\input{sections/solving}

 \section{Experiments}\label{sec:Experiments}
\input{sections/experiments}

\section{Conclusion}
\input{sections/conclusion}

\bibliographystyle{IEEEtran}
\bibliography{references}

\end{document}

%% file: macros.tex
\renewcommand{\vec}[1]{\mathbf{#1}}

\newcommand{\R}{\mathbb{R}}

\newcommand{\cupdot}{\mathbin{\mathaccent\cdot\cup}}

\newcommand{\tup}[1]{\left\langle #1\right\rangle}

\newcommand{\domain}{\mathcal{D}}
\newcommand{\states}[1]{{S_{#1}}}
\newcommand{\initState}{{s_0}}
\newcommand{\actions}[1]{{A_{#1}}}
\newcommand{\labelling}{L}
\newcommand{\ap}{\mathit{AP}}
\newcommand{\strat}[1]{{\pi_{#1}}}
\newcommand{\run}{\rho}

\newcommand{\AP}{\ensuremath{\mathrm{AP}}}
\newcommand{\prop}[1]{\mathtt{#1}}
\newcommand{\nextt}{\xspace\bigcirc\xspace}
\newcommand{\true}{\top}
\newcommand{\globally}{\xspace\square\xspace}
\newcommand{\finally}{\xspace\lozenge\xspace}
\newcommand{\until}{\ U\ }

\newcommand{\game}{\mathcal{G}}
\newcommand{\acc}{\Omega}
\newcommand{\paritygame}[1]{{\mathit{Parity}[#1]}}
\newcommand{\col}{c}
\newcommand{\template}{\Pi}
\newcommand{\unsafe}{U}
\newcommand{\colive}{C}
\newcommand{\livegroups}{\mathcal{H}}
\newcommand{\live}{H}

\newcommand{\avoidRWA}{\ensuremath{\mathcal{A}}}

\newcommand{\turtlebot}{TurtleBot4}

\newcommand\dist{3.5cm}

\newcommand\drawareas{
    \draw[area, draw=blue, fill=blue!20] (0,0) rectangle (10,15) node[pos=.5] (Q3) {};
    \draw[area, draw=green, fill=green!20] (10,0) rectangle (15,15) node[pos=.5] (Q4) {};
    \draw[area, draw=red, fill=red!20] (15,0) rectangle (25,15) node[pos=.5] (Q5) {};
    \draw[area, draw=pink, fill=pink!20] (0,15) rectangle (15,25) node[pos=.5] (Q2) {};
    \draw[area, draw=yellow, fill=yellow!20] (15,15) rectangle (25,25) node[pos=.5] (Q1) {};
    \draw[draw=black, line width=0.5mm] (0,0) rectangle (25,25);
}

\newcommand\drawavoid{
	\draw [pattern={Lines[angle=45, distance=1.5mm, line width=0.2mm]}, pattern color=red] (15,15) rectangle (25,25);
}

\newcommand\drawcbot[1]{\includegraphics[width=#1]{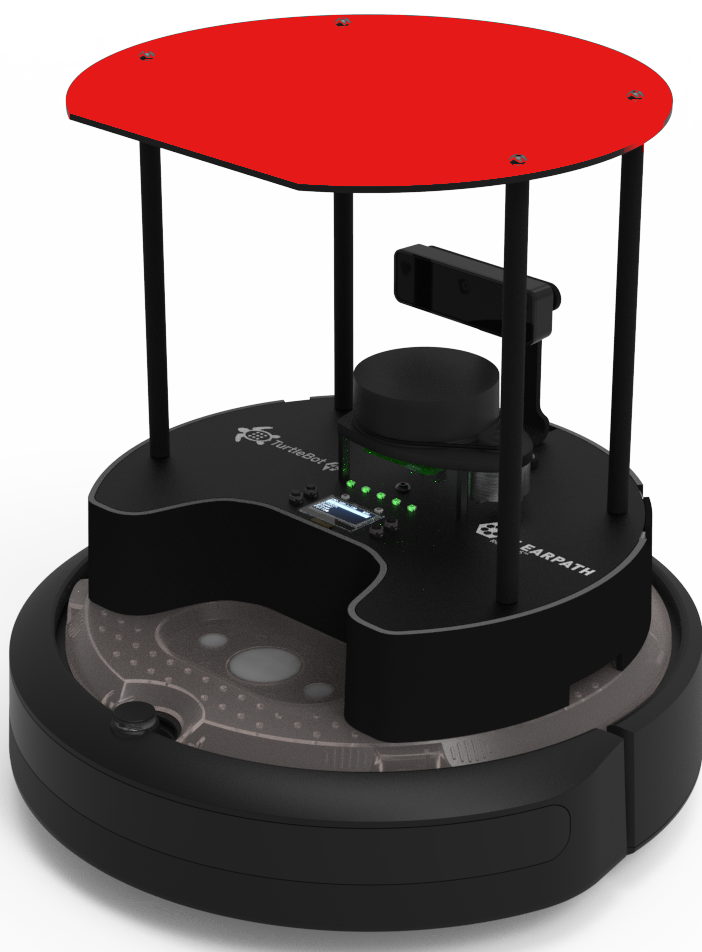}}
\newcommand\drawebot[1]{\includegraphics[width=#1]{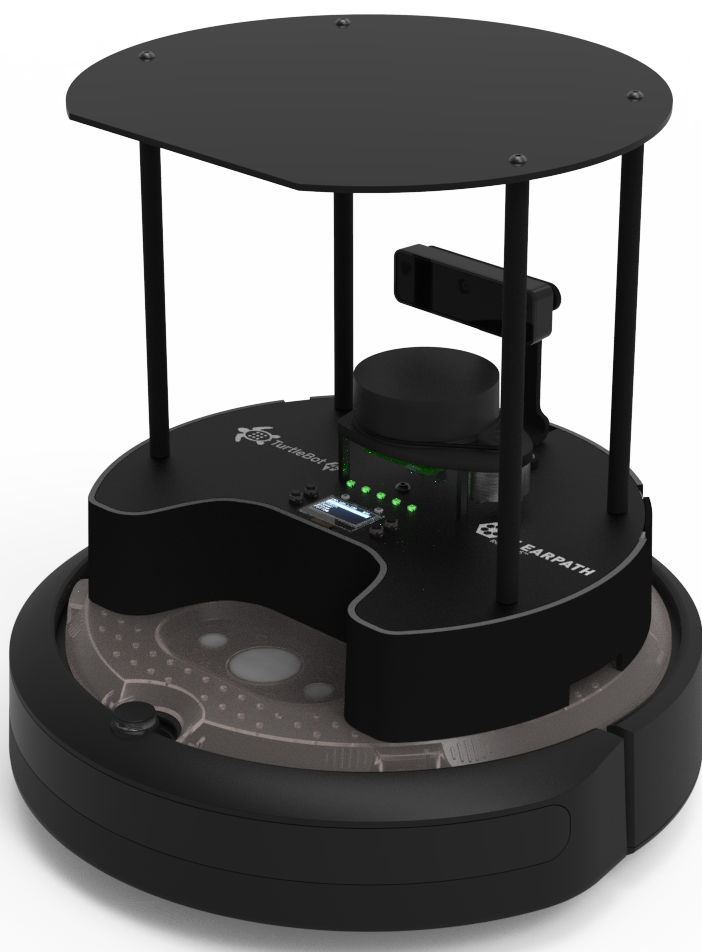}}

%% file: sections/intro.tex
Autonomous systems are becoming increasingly prevalent and already engage in complex interactions among each other and with humans -- such as self-driving cabs entering city traffic or factory robots cooperating with humans to assemble products.
While reliability of autonomous system interaction is crucial in such situations and has received overwhelming attention in recent years \cite{Review23DataDrivenSafetyFiltersAmesTomlinZeilingerEtAl,Review23SafetyFilterHsuHuFisac}, also \emph{performance} is needed to not make complex procedures stall under increasing autonomous participants.

As an example, consider multiple autonomous cars driving on a highway. When only concerned with safety by avoiding collisions with other cars, an agent might miss its highway exit as it is unable to \emph{actively} enforce merging to the right lane in time. Similarly, autonomous 
warehouse robots can be caught in a live-lock dynamically giving right to each other, without enforcing progress towards their destination\footnote{See \url{https://youtu.be/AAhHmquT9Fs}}.

These examples demonstrate that autonomous system interaction requires both (high-level, long-term) \emph{strategic interaction} -- i.e., merging to the right lane when the exit comes closer or always eventually dropping a parcel correctly -- and (low-level, short-term) \emph{dynamic interaction} -- i.e., anticipating how other agents will give way to actually achieve strategic goals. In addition, this behavior should reliably emerge while agents can only \emph{observe} other agents' behavior in real-time and have no access to their goals.

This paper presents a practical solution to this multi-agent control problem via a novel integration of (high-level) \emph{strategic} and (low-level) \emph{dynamic games} into a so-called \emph{context-triggered contingency game} with a real-time capable model predictive control (MPC) implementation. The resulting controller ensures the satisfaction of a reactive strategic specification of the ego agent given in linear temporal logic (LTL) while locally optimizing strategic choices based on the estimated actions of other agents through their dynamic interaction. The resulting efficient and reliable interaction of agents is demonstrated in an autonomous driving simulation and on robotic hardware.

\subsection{Related Work}

Layered control architectures are known to be essential to address the increasing complexity of autonomous systems \cite{matni-2024} while algorithmic game theory \cite{roughgarden2010algorithmic} is known to capture the mutual influence of interacting agents in a concise fashion. While (high-level, long-term) strategic interactions are usually captured by (logical) games on graphs \cite{KressGazitFainekosPappas2009,AlthoffBeltaReviewFMCEforAutonomousDriving}, (low-level, short-term) dynamic interactions are usually captured by (continuous) dynamic games \cite{sadigh2016planning}. Existing work usually integrates \emph{one} such game into one layer of  a multi-layered control architecture and ignores game-like aspects in other layers. To the best of the authors' knowledge, this is the first work which integrates both types of games across layers while still achieving a real-time capable implementation.

\smallskip
\noindent\textit{(High-level) Strategic Games.\footnote{We note that we use the term \emph{strategic game} to refer to a (multi-round) reactive graph game -- different from (one-shot) strategic normal-form games.}}
There exists a large body of work on the use of \emph{strategic games} derived from LTL specifications in higher layers of multi-level controllers, recently reviewed in \cite{YIN2024100940} and discussed in recent textbooks \cite{tabuada-2009-verification,belta2017formal,lindemann2025formal}. Importantly, most solutions consider only a \emph{single agent} which either (i) only interacts with boolean aspects of the environment (i.e., adapts to a door being open/closed, or an alarm being on/off),
or (ii) only executes a non-reactive task plan %
implemented via control barrier functions to allow online adaptation to dynamic uncertainty. %
A recent exception is \cite{nayak-2023} which is based on a new representation of strategies in the form of \emph{strategy templates} \cite{anand-2023-synthesizing}, which are then translated into control Lyapunov functions. While this allows for both boolean interaction with the environment and dynamic adaptations, it is only applicable to a single agent. 

In \emph{multi-agent settings} typically a centralized game is solved and the resulting  action profile is implemented via (decentralized or multi-agent) control barrier functions, see e.g. \cite{9696363,lindemann2025formal,10937075}. %
Crucially, these works pre-compute an action profile given specifications for all players which is then realized dynamically. In contrast, we only consider knowledge of the ego agent's specification and therefore defer large parts of the strategy selection to the online phase when actions of other agents can be inferred from observations. However, in contrast to multi-level planning \& MPC frameworks like \cite{kalluraya-2023-multi,zhang-2023-multi} we guarantee the eventual satisfaction of the ego agent's objective without replanning.

\smallskip
\noindent\textit{(Low-level) Dynamic Games.}
On the other end of the spectrum, dynamic games arise from multi-agent feedback control problems in highly interactive domains such as autonomous driving \cite{fisac-2018-hierarchical} and racing scenarios \cite{spica-2018-real, zhu-2022-sequential}. To practically apply them for feedback control, the resulting nonlinear optimal control problem must be solved in real-time, which induces severe computational challenges.
ALGAMES \cite{lecleach-2021-algames} was the first general-purpose solver capable of handling nonlinear equality and inequality constraints.
While showing good performance, it can struggle with convergence in scenarios with complex dynamics and constraints \cite{zhu-2022-sequential}.
More recently, the dynamic game sequential quadratic programming (DG-SQP) solver was introduced \cite{zhu-2022-sequential}.
It shows improved solution quality compared to ALGAMES, but still suffers from long solving times.

While this makes the solution of dynamic games already computationally intense, the situation becomes even more complicated if uncertainty about the behavior of other agents must be taken into account.
To address this problem, branch MPC \cite{chen-2021-interactive} and topology-driven MPC \cite{degroot-2025-topology} generate multiple independent trajectories and select the most suitable one based on a risk measure. However, these approaches cannot smoothly blend trajectories under high uncertainty. %
To address this, contingency planning approaches such as \cite{mustafa-2024-racp} were introduced, which have recently been extended to \emph{contingency games} \cite{peters-2023-contingency}, which incorporate multiple interactive future scenarios into a single dynamic game.

While \cite{fisac-2018-hierarchical} investigated the multi-level solution of a (classical) dynamic game, we are not aware of any existing integration of contingency games into a multi-level control framework, as presented in this paper. In addition, none of the existing solvers for dynamic games is able to handle the complexity of the resulting optimal control problems in real-time. To address this shortcoming, an important contribution of this paper is the development of a novel factor-graph-based approximate solver for contingency games.

\subsection{Contribution \& Outline}
The key contribution of this paper is the implementation of a real-time capable two-layered reactive planning and control architecture which instantiates and solves a contingency game per strategic game node, leading to a dynamic sequence of contingency games driven by the long-term strategic (LTL) objective of the ego agent (see Fig.~\ref{fig:robots_game} for an illustration). 

Concretely, this paper presents:
\begin{inparaenum}
 \item the integration of strategy templates \cite{anand-2023-synthesizing} and contingency games \cite{peters-2023-contingency} into a multi-layered \emph{context-triggered contingency game}, which is formalized in Sec.~\ref{sec:Games} and illustrated for a robot application scenario in Sec.~\ref{sec:Setup};
 \item the development of a novel factor-graph-based solver which is demonstrated to outperform existing solutions to contingency games in both speed and scalability and which enables a real-time capable MPC formulation of the resulting feedback controller, as discussed in Sec.~\ref{sec:Solver}; and
 \item an experimental validation of the resulting controller on hardware and in simulation presented in Sec.~\ref{sec:Experiments}.
\end{inparaenum}

%% file: sections/problemSetup.tex
We focus on scenarios where multiple agents interact both strategically and physically and each agent has its own objective -- unknown to the others. Throughout this paper, we take the view of an ego agent with a fixed (long-term) strategic objective. %
We consider two application scenarios. 

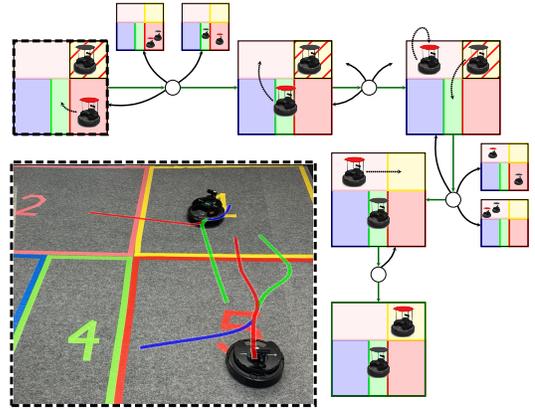
\begin{figure}
\centering
  \resizebox{0.8\columnwidth}{!}{\input{figs/robots-graph}}
  \caption{Mobile Robot Navigation Scenario. Top right: Part of the strategic game graph. Box and circle nodes are owned by environment and ego agent, respectively. Bottom left: Contingency game solution for the dynamic game induced by the top-left context (dashed boundary). Matching colored trajectories illustrate matching intents.  }\vspace{-0.4cm}
  \label{fig:robots_game}
\end{figure}

(1) A \emph{mobile robot} navigating an unstructured environment containing unknown, potentially moving obstacles (e.g., humans, other agents, or furniture) to fulfill temporal tasks in interaction with other agents and humans. This scenario is illustrated in Fig.~\ref{fig:robots_game} and will be used as a running example throughout this paper. The resulting controller was implemented and tested on hardware (see Sec.~\ref{sec:Experiments}).

(2) An \emph{autonomous driving} scenario, where an ego vehicle has to maintain a constant cruising speed while always moving back to the right lane again. This scenario is further described in Sec.~\ref{sec:Experiments} and the resulting controller was implemented and tested in simulation. %

\subsection{Interaction Models}%

\smallskip
\noindent\textit{(High-level) Reactive Planning Domain. } %
We model all relevant strategic interactions between the (controlled) ego agent and (uncontrolled) agents in its environment as a reactive planning domain $\domain = \tup{\states{},\initState,\actions{},\ap_D,\labelling}$, where:
\begin{inparaenum}[(i)]
    \item $\states{} = \states{c}\cupdot \states{u}$ is a set of \emph{contexts}, partitioned into controlled contexts $\states{c}$ owned by the ego agent and uncontrolled contexts $\states{u}$ owned by environment agents;
    \item $\initState \in \states{}$ is the initial context;
    \item $\actions{} = \actions{c}\cupdot \actions{u}$ is the set of actions (modeled as directed edges), partitioned into controlled actions $\actions{c} \subseteq \states{c}\times \states{u}$ and uncontrolled actions $\actions{u} \subseteq \states{u}\times \states{c}$;
    \item $\ap_D$ is a set of boolean variables; %
    \item $\labelling\colon \states{} \to 2^{\ap_D}$ is a labeling function that labels each context with a set of propositions which are true in this context.
\end{inparaenum}

The planning domain can for example be specified using the planning domain description language (PDDL)~\cite{pddl}, a standard language in the field of autonomous systems. %

\smallskip
\noindent\textit{(Low-level) System Dynamics.}
We assume that all agents $p$ can be modeled by continuous-time affine dynamics
\begin{equation}\label{eq:affine-dynamics}
  \dot{x}^p = f^p(x^p) + g^p(x^p)u^p,
\end{equation}
where $X \subseteq \R^{n}$ is the set of admissible physical states, $U \subseteq \R^{m}$ is the set of admissible inputs, and the functions $f^p:\mathbb{R}^{n} \to \mathbb{R}^{n}$ and $g^p:\mathbb{R}^{n} \to \mathbb{R}^{n \times m}$ are locally Lipschitz continuous. We assume the existence of a map $\ell^p: X \rightarrow 2^{\ap_S^p}$ which maps every continuous state of agent $p$ to a set of its system propositions $\ap^p_S$, i.e., the truth value of these propositions is dependent on the current (continuous) state of an agent and can be actively changed by controlling the dynamics of agents.

\begin{example}[Robot Domain and Dynamics]
We consider a simplified reactive planning domain consisting of five regions $Q_{1},…,Q_{5}$ as depicted in Fig.~\ref{fig:robots_game}, a controlled agent $R_{c}$, and an uncontrollable environment agent $R_{u}$. The states of the planning domain model all possible locations of both agents, and their possible actions to leave or enter particular regions induced by the adjacency of workspace regions. %
The \emph{state propositions} are naturally defined for $1 \leq i \leq 5$ s.t.\ $\prop{c}_{i}$, $\prop{u}_{i}$, $\prop{h}_u$ are set to $\prop{true}$ if and only if agent $R_{c}$, agent $R_{u}$ or a human is in region $Q_{i}$, respectively. Then we have a simple map $\ell^p$ s.t.\ $\ell^p(x^p)=\prop{p}_i$ if $x^p\in Q_i$ for $p\in\{u,c\}$.
\end{example}

Depending on the application, the reactive planning domain $\domain$ might need to be refined to truthfully reflect the interaction of $\ap_S$ and $\ap_D$. %
To simplify the presentation, we assume $\ap_S\subseteq \ap_D$ throughout this paper.

\subsection{(High-level) Strategic Games.}

\smallskip
\noindent\textit{Strategic Tasks as LTL Specifications.}\label{sec:problemSetup:ltl}
To express a strategic task for the ego agent, we use \emph{linear temporal logic} (LTL) \cite{BaierKatoen08}, which has been extensively used for strategic reactive planning \cite{kress2018synthesisForRobotsReview}.  
Given a set $\ap_\varphi$ of atomic task propositions, LTL formulas are recursively defined as follows:
\[\varphi ::= \true \mid p \mid \neg \varphi \mid \varphi_1 \land \varphi_2 \mid \nextt \varphi \mid \varphi_1 \until \varphi_2\]
where $p \in \ap_\varphi$ is an atomic proposition; $\neg$ and $\land$ are the boolean operators' negation and conjunction, respectively; and $\nextt$ and $\until$ are the temporal operators 'next' and 'until', respectively.
Other standard operators such as disjunction ($\lor$), implication ($\Rightarrow$), finally ($\finally$), and globally ($\globally$) can be derived from the above operators.
The semantics of LTL formulas are defined over infinite sequences of sets of atomic propositions %
and can be found in standard books~\cite{BaierKatoen08}.

\begin{example}[Robot Specification]
In contrast to state propositions, \emph{task propositions} can additionally model state-independent aspects, e.g.\ that both agents can be requested to \emph{leave} or \emph{enter} a particular region $Q_{i}$ by the other agent or by a human operator. Here, the task proposition $\prop{enter}^e_i$ is true if agent $R_e$ is requested to enter region $Q_i$, with $\prop{leave}^c_i$, $\prop{leave}^e_i$ and $\prop{enter}^c_i$ being defined analogously.

As a simple strategic specification for $R_{c}$ consider the task of always fulfilling an issued enter request for any region, while (i) always avoiding obstacles, (ii) not sharing $Q_1$ with $R_u$, and (iii) not sharing $Q_4$ with a human. This can be formalized by the LTL specification
\begin{subequations}
 \begin{align}\label{equ:example:phi_G}
\varphi_G = &\textstyle\left( \bigwedge_{i=1,…,5} \finally \globally \mathtt{enter}_i^c \Rightarrow \finally \mathtt{c}_i \right)\nonumber\\
&\land \left( \mathtt{u}_1 \Rightarrow \neg \mathtt{c}_1 \right)
\land \left( \mathtt{h}_4 \Rightarrow \neg \mathtt{c}_4 \right).
\end{align}
In addition, we assume that $R_{u}$ respects a leave request for region $Q_1$ only when $R_{c}$ is in region $Q_2$, formalized via
\begin{equation}\label{equ:example:phi_A}
\varphi_A =\left( \prop{leave}_1^e \implies \nextt \neg \prop{u}_1 \right) \land
\left(  \nextt \prop{leave}_1^e \implies \prop{c}_2 \right). %
\end{equation}
\end{subequations}
Combining \eqref{equ:example:phi_G} and \eqref{equ:example:phi_A} yields the strategic objective $\varphi = \globally \varphi_A \Rightarrow \globally \varphi_G$ requiring that the ego agent fulfills $\varphi_G$ when interacting with another agent fulfilling $\varphi_A$. %
\end{example}

As illustrated above, the satisfaction of a specification is usually not under full control of the ego agent, as some propositions cannot be influenced, e.g. whether $R_u$ enters $Q_2$. The strategic cooperation needed to fulfill its specification is captured by a strategic game, as discussed next.

\smallskip
\noindent\textit{Strategic Two-Player Games.}
A (high-level) strategic game is a pair $\game=\tup{\domain,\acc}$, where $\domain = \tup{\states{},\initState,\actions{},\ap,\labelling}$ is a reactive planning domain, and $\acc\subseteq \states{}^\omega$ is a set of infinite sequences of contexts that defines the winning condition of the game. A strategic game is typically constructed as the product of an original planning domain $\domain'$ (e.g.\ derived from PDDL) and an LTL specification $\varphi$ translated into an automaton $\mathcal{A}$ with a winning condition $\acc'$ (see \cite{BaierKatoen08} for details), potentially refined by abstractions of the underlying dynamics \eqref{eq:affine-dynamics}.

The resulting strategic game $\game=\tup{\domain,\acc}$ is played between the ego agent and the environment agents who move a token over the graph, which models their context-dependent strategic interaction. That is, the agent who owns a context chooses an action and therefore determines the next context.

\begin{example}[Strategic Game]
An illustrative part of the game graph for the robot scenario is depicted in  Fig.~\ref{fig:robots_game} (top). The illustration hides the annotation of each context by the current active enter request, which determines which state of the graph needs to be reached by $R_{c}$ to fulfill $\varphi$. %
\end{example}

\smallskip
\noindent\textit{Winning Strategies.}
The solution of a strategic game $\game=\tup{\domain,\acc}$ (formalized in Sec.~\ref{sec:Games}) equips the ego agent with a strategy to choose actions over time.
The problem with the implementation of this solution, however, is that all agents interact highly dynamically. In reality, cars or robots do not patiently wait for others to complete their move such that an action can be taken under full observation. Instead, every agent is moving at the same time and other agents' actions must be anticipated in real-time.
This paper provides a novel solution to this problem by equipping every ego-agent context with a contingency game.

\subsection{(Low-level) Contingency Games}
A contingency game \cite{peters-2023-contingency} is a dynamic game that takes multiple possible intents (i.e., discrete actions) $\Theta^{p}$ of players into account. The solution is obtained by solving a carefully constructed optimal control problem, determining the next input $u_{t+1}$ for the ego agent.
In addition to predicting actions of other players, the formulation allows to include additional low-level objectives, such as collision avoidance, trajectory smoothness or energy consumption limitations.

\begin{example}[Robot Contingency Game]
Consider the top left context in Fig.~\ref{fig:robots_game} where $u_1,c_5, enter_c^1$ are true, i.e., the ego agent is in region $5$ and requested to enter region $1$, which is occupied by the environment agent. As both agents are not allowed to stay in $Q_1$ together, the ego agent has different optimal strategies, depending on the movement of the environment agent. If the environment agent stays in $Q_1$ (blue), $R_c$ should proceed towards $Q_4$ and then $Q_2$ to request $R_u$ to leave $Q_1$ (blue). If the environment agent is however already leaving $Q_2$ (red or green) the ego agent should proceed towards $Q_1$ (red or green), but not enter until the other agent has left.
The resulting contingency game computes beliefs over the intents of the environment agent and computes a weighted blend of the resulting ego-agent trajectories to determine its next control input $u_{t+1}$.
\end{example}

%% file: figs/robots-graph.tex
\begin{tikzpicture}
    \tikzset{auto, >= stealth}
    \tikzset{every edge/.append style={thick, shorten >= 1pt, line width=0.2cm}}
    \tikzset{active/.style={black!60!green}}
    \tikzset{initial/.style={draw, thick, <-, shorten <=1pt}}
    \tikzset{player2/.style = {draw, ultra thick, shape=rectangle, minimum size=5mm}}
    \tikzset{player1/.style = {draw, ultra thick, shape=circle, minimum size=20mm}}
    \tikzset{bplayer0/.style = {draw, thick, shape=ellipse, minimum size=3mm,text width=0.65cm}}
    \tikzset{area/.style = {line width=2mm}}

	\begin{scope}[scale=0.5, shift={(0,0)}, local bounding box=c5e1]
		\drawareas
		\drawavoid
        \draw[black, dashed, dash pattern=on 10mm off 5mm, line width=4mm] (0,0) rectangle (25,25);
        \node (bot1) at (20,6) {\drawcbot{3cm}};
        \node (ebot1) at (19,20) {\drawebot{3cm}};
		\node (target1) at (12.5,9) {};
    \end{scope}
	\node at (c5e1.south west) {\tikzmark{foo}};

    \begin{scope}[scale=0.5, shift={($(c5e1.south east)+(35,0)$)}, local bounding box=c4e1]
		\drawareas
		\drawavoid
        \node (bot2) at (12.5,9) {\drawcbot{3cm}};
        \node (ebot2) at (19,20) {\drawebot{3cm}};
		\node (target2) at (6,20) {};
    \end{scope}
	
    \begin{scope}[scale=0.5, shift={($(c4e1.south east)+(20,0)$)}, local bounding box=c2e1]
		\drawareas
		\drawavoid
        \node (bot3) at (6,20) {\drawcbot{3cm}};
        \node (ebot3) at (19,20) {\drawebot{3cm}};
		\node (target3) at (12.5,9) {};
    \end{scope}

    \begin{scope}[scale=0.5, shift={($(c4e1.south east)+(0,-30)$)}, local bounding box=c2e4]
		\drawareas
        \node (bot4) at (6,20) {\drawcbot{3cm}};
        \node (ebot4) at (12.5,9) {\drawebot{3cm}};
		\node (target4) at (19,20) {};
    \end{scope}

    \begin{scope}[scale=0.5, shift={($(c2e4.south west)+(0,-40)$)}, local bounding box=c1e4]
		\drawareas
        \draw[active, line width=2mm] (0,0) rectangle (25,25);
        \node (bot5) at (19,20) {\drawcbot{3cm}};
        \node (ebot5) at (12.5,9) {\drawebot{3cm}};
    \end{scope}

    \draw[->] (bot1.west) edge[bend left=30, dotted] node[below left] {} (target1);
    \draw[->] (bot2.west) edge[bend left=30, dotted] node[above right] {} (target2);
	\draw[->] (ebot3.west) edge[bend right=30, dotted] node[below left] {} (target3);
    \draw[->] (bot3.west) edge[loop,out=120,in=90,dotted,min distance=4cm] node[above left] {} (bot3.north);
	\draw[->] (bot4.east) edge[dotted] node[above] {} (target4);
	
    \node[player1] (c5e1to4) at ($(c5e1)!0.5!(c4e1)$) {};

    \begin{scope}[scale=0.25, shift={($(c5e1to4)+(-5-25,20)$)}, local bounding box=c5e5]
		\drawareas
        \node (a) at (18,4) {\drawcbot{1cm}};
        \node (b) at (22,8) {\drawebot{1cm}};
    \end{scope}

    \begin{scope}[scale=0.25, shift={($(c5e1to4)+(5,20)$)}, local bounding box=c5e2]
		\drawareas
        \node (c) at (20,6) {\drawcbot{1cm}};
        \node (d) at (12.5,9) {\drawebot{1cm}};
    \end{scope}

    \node[player1] (c4e1to2) at($(c4e1)!0.5!(c2e1)$) {};
    \node[player1] (c2e1req) at($(c2e1.south)+(0,-12.5/2-2.5)$) {};
    \node[player1] (c2e4to1) at($(c2e4)!0.5!(c1e4)$) {};

    \node (c4e1to2a) [above left=\dist of c4e1to2] {};
    \node (c4e1to2b) [above right=\dist of c4e1to2] {};
    \node (c2e4to1a) [left=\dist of c2e4to1] {};
    \node (c2e4to1b) [right=\dist of c2e4to1] {};

    \begin{scope}[scale=0.25, shift={($(c2e1req)+(15,5)$)}, local bounding box=c2e5]
		\drawareas
        \node (a) at (6,20) {\drawcbot{1cm}};
        \node (b) at (20,6) {\drawebot{1cm}};
    \end{scope}
    \begin{scope}[scale=0.25, shift={($(c2e1req)+(15,0-25)$)}, local bounding box=c2e2]
		\drawareas
        \node (a) at (3,18) {\drawcbot{1cm}};
        \node (b) at (8,21) {\drawebot{1cm}};
    \end{scope}

    \draw[->,active] (c5e1) edge (c5e1to4);
    \draw[->] (c5e1to4) edge[bend left=20] (c5e1);
    \draw[->] (c5e1to4) edge[bend left=20] (c5e5);
    \draw[->] (c5e1to4) edge[bend right=20] (c5e2);
    \draw[->,active] (c5e1to4) edge (c4e1);
    \draw[->,active] (c4e1) edge (c4e1to2);
    \draw[->] (c4e1to2) edge[bend left=20] (c4e1);
    \draw[->,active] (c4e1to2) edge (c2e1);
    \draw[->,active] (c2e1) edge (c2e1req);
    \draw[->] (c2e1req) edge[bend left=20] (c2e1);
    \draw[->,active] (c2e1req) edge (c2e4);
    \draw[->,active] (c2e4) edge (c2e4to1);
    \draw[->,active] (c2e4to1) edge (c1e4);
    \draw[->] (c2e1req) edge[bend right=30] (c2e2);
    \draw[->] (c2e1req) edge[bend left=30] (c2e5);
    \draw[->] (c2e4to1) edge[bend right=20] (c2e4);
    \draw[->] (c4e1to2) edge[bend right=20] (c4e1to2a);
    \draw[->] (c4e1to2) edge[bend left=20] (c4e1to2b);

    \node[draw, dashed, dash pattern=on 10mm off 5mm, line width=4mm, inner sep=3mm] (pic) at (20,-20) {\includegraphics[width=40cm]{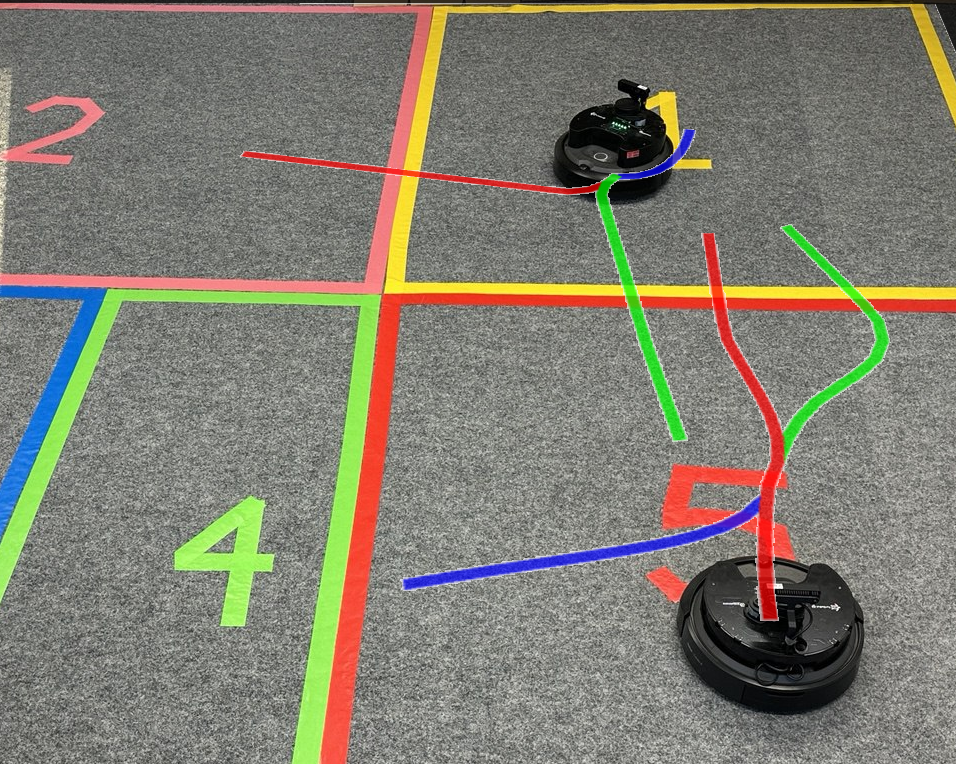}};

\end{tikzpicture}

%% file: sections/combining.tex
Given the problem setup discussed in Sec.~\ref{sec:Setup}, we now formalize the integration of strategic and contingency games. This is done by deferring the choice of strategic actions of the ego agent to the online phase, when observations can be used to infer the current actions of other agents. This is enabled by utilizing a novel, permissive representation of winning strategies for strategic games, called strategy templates \cite{anand-2023-synthesizing}.

\subsection{(High-level) Strategy Templates}
In two-player games, a \emph{strategy template} generalizes the notion of a strategy by succinctly representing an infinite family of strategies through local constraints. %

\smallskip
\noindent\textit{Strategies.}
Let $\game=\tup{\domain,\acc}$ be a strategic game. Then a run $\run = s_0 s_1 s_2 \ldots\in S^\omega$ over $\domain$ is an \emph{infinite} sequence of contexts such that $s_0 = \initState$ and for all $i \geq 0$, $(s_i, s_{i+1}) \in \actions{}$, i.e., there is an action that takes the system from context $s_i$ to context $s_{i+1}$.
A \emph{control strategy} $\strat{c}\colon \states{}^*\states{c} \to \actions{c}$ is a function that maps a sequence of contexts (representing the history of the interaction) to the action that the ego agent will take if it is their turn.
A run $\run = s_0 s_1 s_2 \ldots$ is said to be a $\strat{c}$-run if for all $i \geq 0$, whenever $s_i \in \states{c}$, then $s_{i+1} = \strat{c}(s_0 s_1 \ldots s_i)$.
A control strategy $\strat{c}$ is called \emph{winning} if all $\strat{c}$-runs are contained in $\acc$. The construction of a game $\game$ from an LTL specification $\varphi$ ensures that this implies that all $\strat{c}$-runs satisfy $\varphi$. 

\smallskip
\noindent\textit{Parity Games.}
In this paper we consider the class of \emph{parity games}~\cite{BaierKatoen08} which result from the translation of an LTL formula into a game $\game=\tup{\domain,\acc}$. In a parity game the winning condition $\acc = \paritygame{\col}$ is defined by a coloring function $\col\colon \states{} \to \mathbb{N}$ that assigns a natural number (color) to each state, and a run $\run$ belongs to $\paritygame{\col}$ if the maximum color that appears infinitely often in $\run$ is even. %

\smallskip
\noindent\textit{Strategy Templates.}
Given a strategic game $\game=\tup{\domain,\acc}$, a \emph{strategy template}~\cite{anand-2023-synthesizing} is a tuple
$\Psi = (\unsafe, \colive, \livegroups)$ consisting of three disjoint sets of actions controlled by the ego agent:
\begin{inparaenum}[(i)]
    \item \emph{Unsafe actions} $\unsafe\subseteq\actions{i}$: actions that the agent is prohibited from taking;
    \item \emph{Co-live actions} $\colive\subseteq\actions{i}$: actions that may only be taken finitely many times along any run;
    \item \emph{Live-groups} $\livegroups\subseteq 2^\actions{i}$: sets of actions such that, if the source state of some $\live\in\livegroups$ is visited infinitely often, the agent must take at least one action from the set infinitely often.
\end{inparaenum}
A run $\run$ is said to comply with a strategy template $\template$, 
if it satisfies all the specified constraints.
A strategy $\strat{}$ is said to follow a strategy template $\template$, denoted $\strat{} \models \template$, if all $\strat{}$-runs comply with $\template$.
A strategy template $\template$ is called winning if all strategies that follow it are winning.
For a detailed formal definition, further intuition and an algorithm to compute a winning strategy template see~\cite{anand-2023-synthesizing}.

\subsection{(Low-level) Contingency Games}
We assume a set of agents $P$ where each agent $p\in P$ has a fixed set of intents $\Theta^{p}$. We assume that for each agent $p$ and for each intent $\theta \in \Theta^{p}$ its states evolve according to the dynamic constraint
$x_{\theta,t+1}^{p} = f^{p}(x_{\theta,t}^{p}, u_{\theta,t}^{p})$. For simplicity, we assume the state dimensions $n^{p}_{\theta}$ and input dimensions $m^{p}_{\theta}$ are equal to $n$ and $m$ for all players and intents. For ease of notation, we use the shorthand notations
$\vec{x}^{p}_{\theta} = (x_{\theta,t}^{p})_{t=1}^{T} \text{ and } \vec{u}^{p}_{\theta} = (u_{\theta,t}^{p})_{t=1}^{T}$
to denote the states and inputs of player $p$ for intent $\theta$ aggregated over time, as well as $\vec{x}_{\theta} = (\vec{x}^{p}_{\theta})_{p=1}^{N} \text{ and } \vec{u}_{\theta} = (\vec{u}^{p}_{\theta})_{p=1}^{N}$ to denote the stacked state and input vectors of all players. In addition, the shorthands
$\vec{x}_{\Theta} = \left(\vec{x}_{\theta}\right)_{\theta \in \Theta}$ and $\vec{u}_{\Theta} = \left(\vec{u}_{\theta}\right)_{\theta \in \Theta}$
are used to denote the states and inputs aggregated over all hypotheses.

\smallskip
\noindent\textit{Dynamic Objectives.}
We assume the dynamics and intents of players to be coupled such that the objective function $J^{p}(\vec{x},\vec{u}^{p})$ of an agent $p$ might depend on the states of other players, but only on the inputs of $p$.
We define $g^{p}(\vec{x}^{p},\vec{u}^{p})$ and $h^{p}(\vec{x}^{p},\vec{u}^{p})$ to be the equality and inequality constraints of player $p$, respectively, which model \emph{private constraints} that only depend on the states and inputs of player $p$, and include the dynamics $f^{p}$ in the equality constraints.
Additionally, we define $\bar{g}(\vec{x},\vec{u})$ and $\bar{h}(\vec{x},\vec{u})$ to be the \emph{shared constraints} that depend on the states and inputs of all players. %

\smallskip
\noindent\textit{Contingency Games.}
A contingency game \cite{peters-2023-contingency} is a dynamic game that takes multiple possible intents of all players into account, which must be inferred by the ego agent via observations.  
This uncertainty is represented by a discrete probability distribution $b(\theta)$ for $\theta \in \Theta$.
The joint hypothesis space for all players is defined as $\Theta = \Theta^{1} \times \Theta^{2} \times … \times \Theta^{N}$.
The intents of all players are assumed to be independent, hence the belief $b(\theta)$ can be written as the product of the individual beliefs $b(\theta) = \prod_{p=1}^{N} b(\theta^{p}).$

For environment agents, the formulation of the control problem becomes a classical dynamic game per intent $\theta$ formulated as the coupled optimal control problem
\begin{mini}
  {\vec{x}^p_{\theta}, \vec{u}^p_{\theta}}
  {J^p_{\theta} \left(\vec{x}_{\theta}, \vec{u}_{\theta}^p\right)}
  {}
  {}
  \addConstraint{g^p_{\theta}(\vec{x}^p_{\theta},\vec{u}^p_{\theta})}{=0}
  \addConstraint{h^p_{\theta}(\vec{x}^p_{\theta},\vec{u}^p_{\theta})}{\leq 0}  \addConstraint{\bar{g}_{\theta}(\vec{x}_{\theta},\vec{u}_{\theta})}{= 0}  \addConstraint{\bar{h}_{\theta}(\vec{x}_{\theta},\vec{u}_{\theta})}{\leq 0}.\label{equ:op:uc}
\end{mini}
For the ego agent, however, all possible intents are taken into account simultaneously.
This is achieved by modifying the objective function to optimize the expected value over all hypotheses, and by duplicating all constraints per intent:
\begin{mini!}
  {\vec{x}^c_{\Theta}, \vec{u}^c_{\Theta}}
  {\sum_{\theta \in \Theta} b(\theta) J^c_{\theta} \left(\vec{x}_{\theta}, \vec{u}_{\theta}^c\right)}
  {}
  {}
  \addConstraint{g^c_{\theta}(\vec{x}^c_{\theta},\vec{u}^c_{\theta})}{=0}{\quad \forall \theta \in \Theta}
  \addConstraint{h^c_{\theta}(\vec{x}^c_{\theta},\vec{u}^c_{\theta})}{\leq 0}{\quad \forall \theta \in \Theta}
  \addConstraint{\bar{g}_{\theta}(\vec{x}_{\theta},\vec{u}_{\theta})}{= 0}{\quad \forall \theta \in \Theta}
  \addConstraint{\bar{h}_{\theta}(\vec{x}_{\theta},\vec{u}_{\theta})}{\leq 0}{\quad \forall \theta \in \Theta}
  \addConstraint{u^c_{\theta_1,t}}{= u^c_{\theta_2,t}}{\quad \forall \theta_1,\theta_2 \in \Theta}{\nonumber}
    \addConstraint{}{}{\quad \forall t = 1,…,t_b}{\label{prob:contingency:constr:contingency}}
\end{mini!}
The last line \eqref{prob:contingency:constr:contingency} is a \emph{contingency constraint} which ensures that all control inputs up to the \emph{branching time} $t_{b}$ are equal and influenced by all possible contingencies, weighted by the belief for each scenario. After $t_{b}$, it is assumed that certainty is obtained about the intent of the other agents, i.e., there exists an intent $\theta$ with $b(\theta) = 1$.
We refer to \cite{peters-2023-contingency} for a discussion of the branching time, its computation heuristics, its influence on contingency game solutions.   %

As in classical optimal control, the solution of \eqref{equ:op:uc} determines the trajectory of each player $p$ consisting of a sequence of states $x_{1}^{p},x_{2}^{p},\dots,x_{T}^{p} \in \R^{n}$ and control inputs $u_{1}^{p},u_{2}^{p},\dots,u_{T-1}^{p} \in \R^{m}$, if $b(\theta)$ is given. To implement this solution in a model predictive fashion, $b(\theta)$ needs to be estimated from predicted trajectories and $u_1^c$ is used as the next control input for the ego agent in every time step. We refer the reader to \cite{peters-2023-contingency} for more details.

\subsection{Extracting Context-Dependent Intents}\label{sec:CGa}
In order to connect strategic games to contingency games, we now discuss how intents for all agents can be extracted from a particular context $s$ (i.e., a particular node in Fig.~\ref{fig:robots_game}) to formulate a contingency game for this context $s$. 

\smallskip
\noindent\textit{Environment Agents.}
We discussed in Sec.~\ref{sec:Setup} that \emph{state propositions} are directly connected to the low-level dynamics of agents, as their truth value is determined by their (controlled) continuous behavior. Given a strategic game $\game=\tup{\domain,\acc}$ and the current context $s$, we therefore obtain the possible intents of an environment agent $p$ by extracting
\[
  \Theta^{p}_s := \left\{ L(s') \cap \AP_{S}^{p} ~|~ (s,s') \in A_u \right\}
\]
from $\domain$, yielding a context-dependent set $\Theta_s := \times_{p\in E} \Theta^{p}_s$.

\smallskip
\noindent\textit{Ego Agent.}
Given a context $s$ and a specific environment agent intent $\theta \in \Theta_s$, we observe that the latter is connected to an action $(s,s') \in A_u$ where $s'\in S_c$ is an ego agent context. Given that we have pre-computed a strategy template $\Psi$ over $\game$, we can directly extract $\Psi|_{s'} = (\unsafe|_{s'}, \colive|_{s'}, \live|_{s'})$ as the unsafe, colive and live actions originating in $s'$. %

We now utilize $\Psi|_{s'}$ to derive a context- and intent-dependent objective function $J^c_{s,\theta} \left(\vec{x}_{\theta}, \vec{u}_{\theta}^c\right)$ to be used in the context-dependent version of (4). Towards this end, we observe that live actions are those which allow the ego agent to make progress towards the satisfaction of its (high-level) specification, and should therefore be preferred. We therefore choose a goal state $x_{goal}$ for each live action $(s',s'')\in \live|_{s'}$ if $\live|_{s'}\neq\emptyset$, s.t.\ 
$x_{goal}\in {\ell^c}^{-1}(\alpha)$ where $\alpha\in (L(s'')\setminus L(s'))\cap \ap^c_S$, i.e.,  
$x_{goal}$ is a representative goal state (e.g., the center) of the state space region associated with the state proposition $\alpha$ which must become true in $s''$ but is not yet true in $s'$. If $\live|_{s'}=\emptyset$, we choose an edge $(s',s'')\in A_c\setminus \colive|_{s'}\setminus \unsafe|_{s'}$ instead, which always exists.
Given $x_{goal}$, the objective function $J^c_{s,\theta}$ follows straightforwardly.

\subsection{Extracting Context-Dependent Constraints}\label{sec:CGb}
While the previous discussion shows how strategy templates allow us to extract local dynamic goals from the strategic game that enable progress towards satisfying a (long-term) objective, it is equally important to ensure that no (context-triggered) safety violation (as specified in $\varphi$) occurs. In our example, such safety violations occured if a region was shared with other agents, i.e.\ a safety constraint which was only active in a context where another agent was present.

We again use the local strategy template $\Psi|_{s'}$ to include such context-dependent avoid constraints into the contingency game. Towards this goal we define $\kappa:=L(s')$ as the current context and $\mathcal{A}:=\bigcup_{(s',s'')\in\colive|_{s'}\cup \unsafe|_{s'}}L(s'')\cap \ap_S$ as the current propositional avoid set and utilize control barrier functions to formalize their dynamic effect.

\smallskip
\noindent\textit{Control barrier functions (CBFs)} \cite{ames-2019} are a popular tool for guaranteeing safety properties in dynamical systems.
Typically, CBFs are formulated for continuous time dynamics, and are utilized in the form of safety filters to keep the physical state of a control system inside a safe set. To formalize this notion, consider a set $\mathcal{C}$ defined as the superlevel set of a continuously differentiable function $h: D \subset \mathbb{R}^n \to \mathbb{R}$, i.e.,
$\mathcal{C} = \{x \in D : h(x) \geq 0\}$.
Here, safety is defined w.r.t.\ forward invariance, i.e., when starting inside the safe set $\mathcal{C}$, the system is guaranteed to stay in $\mathcal{C}$.

In order to define context-dependent avoid constraints we translate both the context $\kappa$ and the avoid set $\mathcal{A}$ into sets $\mathcal{C}^\kappa\subseteq \ell^{-1}(\kappa)$ and $\mathcal{C}^\mathcal{A}\subseteq \ell^{-1}(\mathcal{A})$ with associated CBFs $h^\kappa$ and $h^\mathcal{A}$. In particular, if each proposition $\alpha \in \avoidRWA$ corresponds to some CBF $h^{\alpha}$, and each proposition $\gamma \in \kappa$ corresponds to a CBF $h^{\gamma}$, then $h^{\avoidRWA}$ and $h^{\kappa}$ can be defined as
  $h^{\avoidRWA} = \bigwedge_{\alpha \in \avoidRWA} h^{\alpha}$ and $ h^{\kappa} = \bigwedge_{\gamma \in \kappa} h^{\gamma}$.

\smallskip
\noindent\textit{Context-Dependent Optimization Constraints.}
Referring back to the intuition that avoid constraints should only hold in a given context, we formalize context-dependent avoid constraints as a Boolean combination of CBFs \cite{glotfelter-2018} s.t.\
\begin{equation*}
\left[ h^{\kappa}(x) \Rightarrow \neg h^{\avoidRWA}(x) \right] = \max(-h^{\kappa}(x), -h^{\avoidRWA}(x)).
\end{equation*}
This formulation allows to model the situation that the context changes externally (i.e.,  a human enters the room) and new avoid constraints are activated.
In addition, it is important to note that some constraints may conflict with the current target. To address this, avoid constraints are only activated when the belief for the corresponding context exceeds a specified threshold. %
From a practical point of view, the outlined procedure to extract context-dependent constraints results in a large number of often redundant functions $h_\theta$.
We therefore simplify the resulting Boolean expressions before solving a contingency game. %
Simplifying the avoid objectives reduces the number of constraints in the resulting optimization problem, while the combination of identical future behavior minimizes the number of intents. %

\subsection{Context-Triggered Contingency Games.}
Combining the above constructions, we get a multi-level game, where at every time step $t\in \mathbb{N}$ the context $s(t)$ and state $x^p(t)$ may change for every agent $p$. The current context $s(t)$ is influenced in real-time by external inputs (i.e., enter requests issued by a human operator), the movement of environment agents (i.e., entering or leaving a specific region), and the movement of the ego agent (i.e., entering or leaving a specific region). The strategic game graph along with the computed strategy templates ensures that the logical dependencies of these context changes w.r.t.\ the given high-level objective $\varphi$ induce an optimal control problem at the current time step $t$ (as discussed in Sec.~\ref{sec:CGa} and Sec.~\ref{sec:CGb}), such that the behavior emerging from the real-time solution of this game and its MPC implementation indeed satisfies the ego agent's objective $\varphi$. Note that this implies that one might solve a different game in every time step.

%% file: sections/solving.tex
\begin{table*}[t!]
  \centering

  \resizebox{0.8\textwidth}{!}{
  \begin{tabular}{ccccccc}
   &&&&&&\\
    \hline
    && $N=10$ & $N=15$ & $N=20$ & $N=25$ & $N=30$  \\
    \hline
    DG-SQP & \makecell{mean/std (s)} & $0.017 \pm 0.005$ & $0.060 \pm 0.012$ & $0.275 \pm 0.192$ & $0.485 \pm 0.159$ & $0.977 \pm 1.143$ \\
    ALGAMES & \makecell{mean/std (s)} & \makecell{ $0.033 \pm 0.004$} &\makecell{ $0.133 \pm 0.009$} &\makecell{ $0.218 \pm 0.030$} &\makecell{ $0.300 \pm 0.052$} &\makecell{ $0.408 \pm 0.020$} \\
    DG-FG & \makecell{mean/std (s)} & $0.003 \pm 0.002$ & $0.015 \pm 0.006$ & $0.022 \pm 0.014$ & $0.027 \pm 0.011$ & $0.039 \pm 0.019$ \\
    \hline\\
  \end{tabular}
  }
  \vspace{0.2cm}
    \centering
  \resizebox{0.8\textwidth}{!}{
  \begin{tabular}{ccccccc}
     \hline
    && $N=10$ & $N=15$ & $N=20$ & $N=25$ & $N=30$ \\
     \hline
    PATH & \makecell{mean/std (s)} & \makecell{ $0.027 \pm 0.227$} &\makecell{ $0.066 \pm 0.245$} &\makecell{ $0.283 \pm 1.321$} &\makecell{ $1.184 \pm 3.659$} &\makecell{ $1.744 \pm 4.764$} \\
    DG-SQP & \makecell{mean/std (s)} & \makecell{$0.105 \pm 0.092$} & \makecell{$0.432 \pm 0.380$} & \makecell{$0.870 \pm 0.547$} & \makecell{$1.460 \pm 0.994$} & \makecell{$1.905 \pm 1.031$} \\
    DG-FG & \makecell{mean/std (s)} & \makecell{$0.001 \pm 0.001$} & \makecell{$0.002 \pm 0.003$} & \makecell{$0.004 \pm 0.005$} & \makecell{$0.006 \pm 0.005$} & \makecell{$0.008 \pm 0.008$} \\
     \hline
  \end{tabular}
  }
    \caption{Solve times (mean and standard deviation in seconds) for lane merge (top) and crosswalk (bottom) scenarios. }
  \label{tab:crosswalk}\vspace{-0.2cm}

\end{table*}

The solution to a dynamic game constitutes a Nash equilibrium (NE), where no player can improve the cost of their strategy without another player having to increase their cost.
Here, we are interested in finding a generalized Nash equilibrium (GNE) which generalizes NEs by allowing each player's strategy to depend on the strategies of other players.
This models the interaction of players in the control problem, and allows anticipating the reaction of other players to own actions. Note that this assumes that players act rationally, but non-cooperatively. Hence, solving contingency games requires finding the solution to a Generalized Nash Equilibrium Problem (GNEP).

\subsection{Existing Solvers}
Most existing solvers approach this problem by searching for local equilibria only, by first deriving Karush-Kuhn-Tucker (KKT) conditions and rewriting them into a general Mixed Complementarity Problem (MCP). The PATH solver \cite{dirkse-1995-path} remains the de-facto standard solver for MCP problems.
Despite its effectiveness, it was not originally designed with real-time performance in mind, and is not able to provide solutions at high frequency.
ALGAMES~\cite{lecleach-2021-algames} is an augmented Lagrangian based solver specifically designed for dynamic game applications, yet it does not scale well with the number of players as required for contingency games.
More recently, DG-SQP \cite{zhu-2022-sequential} was proposed and demonstrated improved solution quality and robustness compared to both PATH and ALGAMES, particularly for challenging racing scenarios.
However, it is still limited by the method's computational complexity and relatively high solving times, making it unsuitable for real-time applications.

\subsection{A Novel Factor Graph Based Solver}
As an alternative to the derivation of an MCP from the KKT conditions, our novel \emph{Dynamic Game Factor Graph} (DG-FG) solver for contingency games utilizes penalty functions to transform the (GNEP) with inequality constraints into an unconstrained, though non-differentiable, optimization problem.
This approach has been previously studied in \cite{facchinei-2010-penalty}, where penalty-based reformulations of GNEPs were shown to provide globally convergent solution methods.
The advantage compared to the approach based on the KKT conditions is that no additional variables corresponding to the Lagrange multipliers need to be introduced, which drastically decreases the problem size. Although penalty methods result in nonsmooth terms, they still work well in practice and can be solved efficiently with an appropriate solver.

DG-FG is based on a direct model of the dynamic game by a factor graph, similarly to \cite{king-smith-2022-simultaneous}, with the following main advantages: First, the factorization naturally induces sparsity, allowing efficient optimization by considering only the local neighborhoods of each variable.
Second, algorithms for solving factor graphs work incrementally, allowing online computation even for large factor graphs.

DG-FG is implemented using SymForce~\cite{martiros-2022-symforce}, a solver for factor graphs using the Levenberg-Marquardt (LM) algorithm, which computes derivatives symbolically to exploit sparsity and common subterms, and generates optimized C++ code suitable for real-time execution.

\subsection{Benchmarks}
\begin{figure}
   \centering
   \begin{subfigure}[b]{0.49\columnwidth}
    \includegraphics[width=\textwidth]{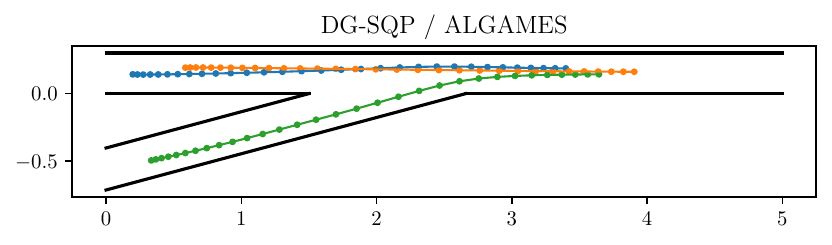}
  \end{subfigure}
  \begin{subfigure}[b]{0.49\columnwidth}
    \includegraphics[width=\textwidth]{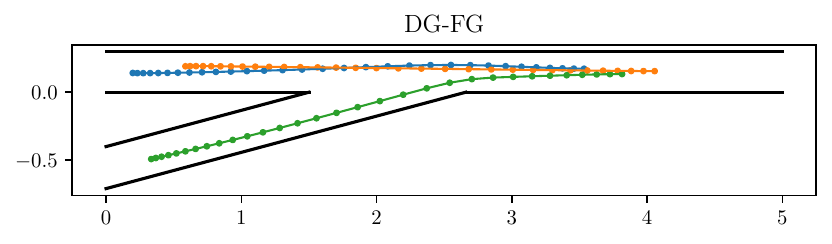}
  \end{subfigure}
   \begin{subfigure}[b]{0.49\columnwidth}
    \includegraphics[width=\linewidth]{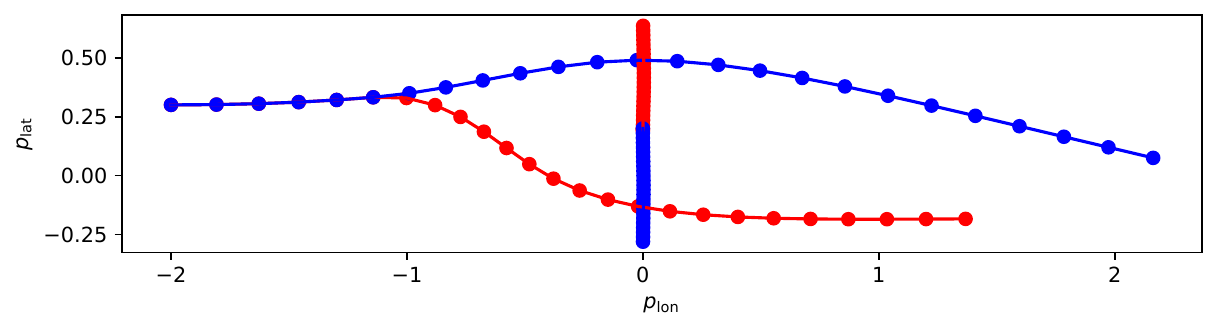}
  \end{subfigure}
  \begin{subfigure}[b]{0.49\columnwidth}
    \includegraphics[width=\linewidth]{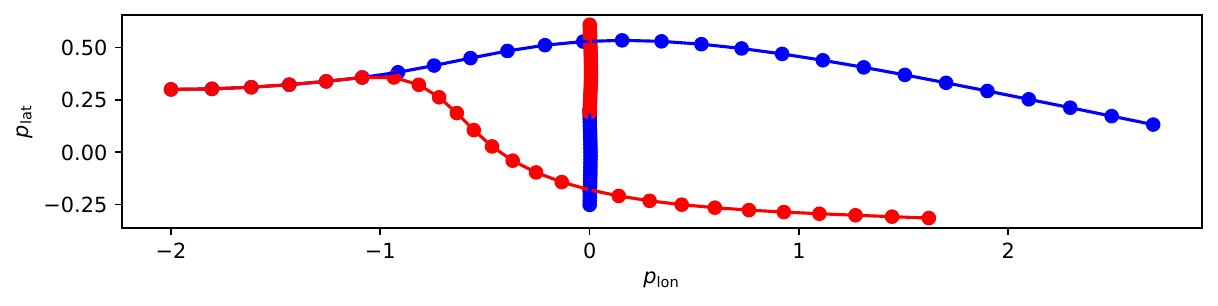}
  \end{subfigure}
  \caption {Solution trajectories for the lane merge (top) and crosswalk (bottom) scenarios. DG-FG solution (right) differs slightly from the reference (left), but remains very similar.}
  \vspace{-0.3cm}
  \label{fig:crosswalk}
\end{figure}

We evaluate DG-FG by comparing its performance to DG-SQP, ALGAMES, and PATH using a ramp merge scenario with three cars from \cite{lecleach-2021-algames} and a crosswalk scenario with one pedestrian and one car from \cite{peters-2023-contingency}, all modelled using nonlinear dynamics.
Performance is measured by solve time and assessed in a Monte Carlo study in which the initial conditions of each agent are randomly perturbed around nominal values.
Each Monte Carlo sample consists of five sequential steps, enabling the solver to warm-start from the previous solution. This procedure provides a more realistic assessment of solver performance in dynamic games.

\cref{fig:crosswalk} shows solutions for two different initial conditions for both scenarios. In both cases, DG-FG trajectories deviate only slightly from the reference and remain qualitatively similar. The similarity can be improved by fine-tuning the cost function weights. Furthermore, \cref{tab:crosswalk} lists solve times for both benchmarks. %
DG-FG provides the fastest solve times by a large margin in all instances.

%% file: sections/experiments.tex
\subsection{Robot Navigation Scenario}
We first consider the robot navigation example depicted in Fig.~\ref{fig:robots_game} and discussed as a running example in Sec.~\ref{sec:Setup}. 
The resulting controller is validated\footnote{Recordings of the experiments are available as supplementary material.} on a \turtlebot~platform running ROS2 jazzy. The DG-FG solver is invoked in real-time with a frequency of 20 Hz, resulting in a smooth trajectory in the presence of multiple simultaneous intents.

\begin{figure}
\begin{center}
   \begin{subfigure}[b]{0.6\columnwidth}
    \includegraphics[width=\textwidth]{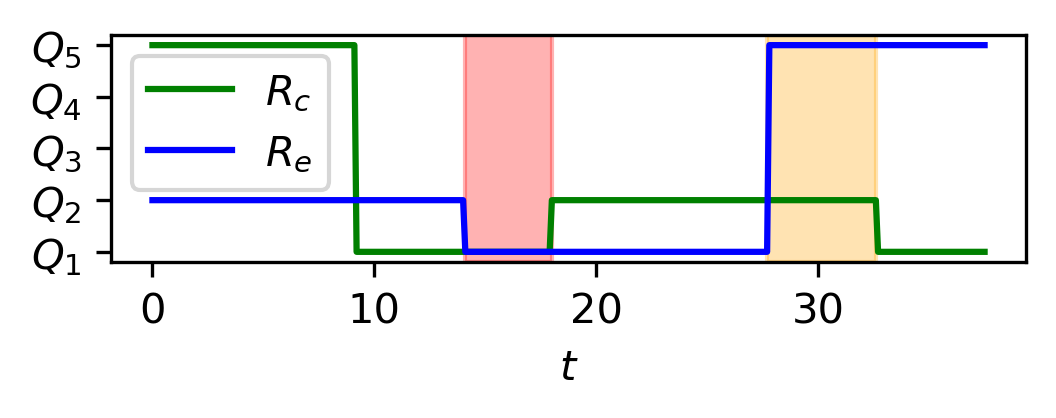}
  \end{subfigure}
  \begin{subfigure}[b]{0.6\columnwidth}
    \includegraphics[width=\textwidth]{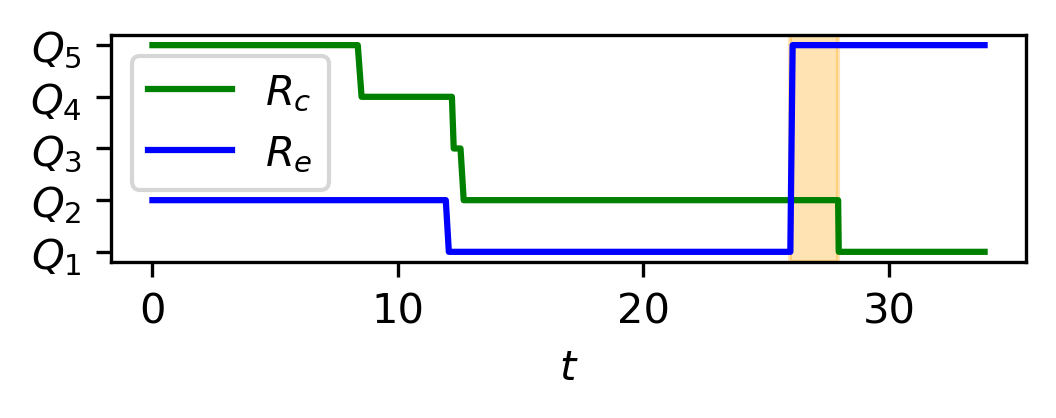}
  \end{subfigure}
  \vspace{-0.2cm}
  \caption{Comparison of the region heat-map resulting from experimental trajectories when $R_c$ is running a context-triggered controller with (bottom) and without (top) contingency games. Specification violation is indicated in red (top). }
  \label{fig:robots_results}
  \vspace{-0.3cm}
\end{center}
\end{figure}

We compare the performance of the resulting controller to a baseline where no contingency game is invoked in the lower level in Fig.~\ref{fig:robots_results} via the following scenario:
Initially, $R_{u}$ is in $Q_2$, while $R_{c}$ is in $Q_5$ and is requested to enter $Q_1$. After $12$ seconds $R_{u}$ enters $Q_1$ and leaves again after $25$ seconds.  We recorded the corresponding trajectories on the testbed and show the resulting region heat map for each agent in Fig.~\ref{fig:robots_results} without (top) and with (bottom) the integration of contingency games. If no contingency game is used, $R_{c}$ directly moves into $Q_1$, causing a violation of the constraint that $R_{u}$ and $R_c$ should never be together in $Q_1$ as soon as $R_{u}$ enters $Q_1$ (red region in Fig.~\ref{fig:robots_results} (top)). This is prevented by invoking a contingency game, as  $R_c$ now anticipates that $R_{u}$ attempts to move into $Q_1$ and therefore directly approaches $Q_2$. We also see that the contingency game solution results in $R_c$ reaching $Q_1$ faster after $R_{u}$ has left (orange region in Fig.~\ref{fig:robots_results}) as it already moves towards $Q_1$ when anticipating that $R_{u}$ will be leaving $Q_1$.

\subsection{Highway Scenario}

This scenario consists of a highway with three lanes.
The objective of the controlled autonomous car (ego agent) is to keep its desired cruising speed, while safely overtaking slower vehicles when necessary.
In doing so, it must comply with traffic rules, avoid collisions, and maintain a safe distance from other vehicles. For simplicity, the highway is assumed to be completely straight, meaning all vehicles move only along the $y$-direction.
Furthermore, it is assumed that all other vehicles also follow the traffic rules. %

The  ego agent's objectives and traffic rules were specified in LTL and translated into a strategic game. The dynamics of each agent are modeled using the nonlinear kinematic bicycle model. The resulting context-triggered contingency game is validated in simulation.
The scenario involves three cars: the car in front of the ego vehicle is slow, while a fast car is on lane 3 (\cref{fig:highway:scenario2:0}).
After observing that the fast car intends to merge into lane 2 (\cref{fig:highway:scenario2:1}), the ego vehicle delays its overtaking maneuver and waits until the fast car has passed (\cref{fig:highway:scenario2:2}).
During this period, it reduces speed to avoid a collision.
After switching to lane 2, it is observed that the car in front is also slowing down, so the ego vehicle switches to lane 3 to safely overtake both cars (\cref{fig:highway:scenario2:4}-g).

\begin{figure}
  \begin{subfigure}[b]{0.49\columnwidth}
    \includegraphics[angle=-90,width=\textwidth]{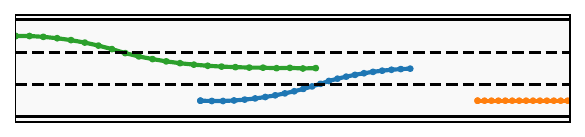}
    \vspace{-0.2cm}
    \caption{MPC; $N=10$}
  \end{subfigure}
  \begin{subfigure}[b]{0.49\columnwidth}
    \includegraphics[angle=-90,width=\textwidth]{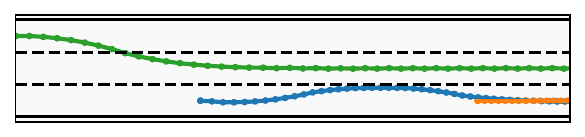}
     \vspace{-0.2cm}
    \caption{MPC; $N=25$}
  \end{subfigure}
  \begin{subfigure}[b]{0.49\columnwidth}
    \includegraphics[angle=-90,width=\textwidth]{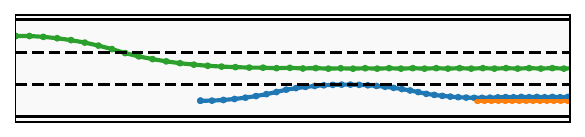}
      \vspace{-0.2cm}
    \caption{Contingency game; $N=10$}
  \end{subfigure}
  \begin{subfigure}[b]{0.49\columnwidth}
    \includegraphics[angle=-90,width=\textwidth]{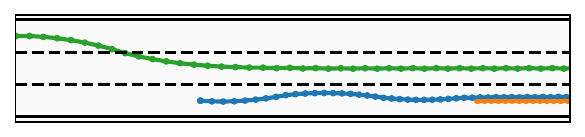}
      \vspace{-0.2cm}
    \caption{Contingency game; $N=25$}
  \end{subfigure}
  \caption{Comparison of reactive and predictive controllers with varying horizon lengths.}
  \label{fig:highway-comparison}
\end{figure}

We compare the performance of our control framework to invoking a classical MPC that only depends on the current context in the lower level as a baseline.
\cref{fig:highway-comparison} illustrates the resulting trajectories of different low-level controllers with two different horizon lengths during the initial part of the scenario.
In both cases, the MPC fails to anticipate the intent of the other driver.
While the MPC with long horizon is still able to take evasive action, the MPC with short horizon merges in front of the other car, forcing it to brake risking a collision.
In contrast, with contingency games, the intent of the other vehicle is correctly anticipated (after it switched lanes), causing the ego vehicle to proactively remain in its lane.
Notably, even with a short horizon, the controller behaves similarly to the MPC with a long horizon.

\begin{figure}
  \centering

  \foreach \i in {0,...,6}{
    \begin{subfigure}[b]{0.1\columnwidth}
      \includegraphics[width=\textwidth]{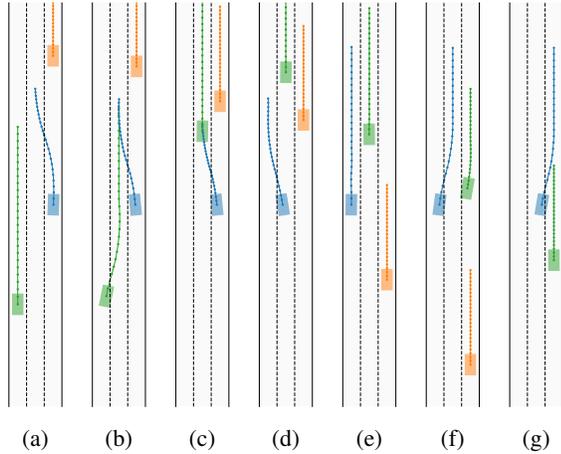}
      \caption{}
      \label{fig:highway:scenario2:\i}
    \end{subfigure}
  }
  \caption{Overtaking maneuver with three vehicles.} %
  \label{fig:highway:scenario2}
  \vspace{-0.3cm}
\end{figure}

%% file: sections/conclusion.tex
We presented context-triggered contingency games, a two-layered architecture combining strategic and dynamic interactions by extracting the intents of each agent from strategy templates.
Our DG-FG solver outperforms existing solvers, enabling safe real-time planning under LTL in interactive scenarios. We validate the approach through hardware and simulation experiments, showing clear improvements over classical MPC baselines.
Future work includes accounting for uncertainties in the dynamics and objectives of other agents, investigating more gradual intent representations for human-robot interaction, and scaling the architecture to more complex multi-agent scenarios.